%% file: main.tex
\documentclass[]{antgroup}
\PassOptionsToPackage{numbers, compress}{natbib}
\usepackage{antgroup}

\usepackage{graphicx}
\usepackage{tikz}
\usepackage{todonotes}
\usepackage{multirow}
\usepackage{amsmath}
\usepackage{cleveref}
\usepackage{subcaption}
\usepackage{multicol}
\usepackage{amssymb}
\usepackage{array}
\usepackage{bm}
\usepackage{enumitem}
\usepackage{algorithm}
\usepackage{algpseudocode}
\usepackage{tabularx}
\usepackage{bbm}
\usepackage{makecell}
\usepackage{manyfoot}
\usepackage{adjustbox}
\usepackage{colortbl}
\usepackage[normalem]{ulem}
\useunder{\uline}{\ul}{}

\input{math_commands}

\input{setting-ai}

\usepackage{xcolor}
\definecolor{firstcolor}{HTML}{C3423F}
\definecolor{secondcolor}{HTML}{2A4B8C}

\title{LLaDA-MoE: A Sparse MoE Diffusion Language Model}

\author{Fengqi Zhu$^{1, 2, *}$, Zebin You$^{1, 2, *}$, Yipeng Xing$^{2, *}$, Zenan Huang$^{2, *}$, Lin Liu$^{2, *}$, Yihong Zhuang$^{2, *}$, Guoshan Lu$^{2, *}$, Kangyu Wang$^{2, 3}$, Xudong Wang$^{2}$, Lanning Wei$^{2}$, Hongrui Guo$^{2}$, Jiaqi Hu$^{2, 4}$, Wentao Ye$^{2, 4}$, Tieyuan Chen$^{2, 3}$, Chenchen Li$^{2}$, Chengfu Tang$^{2}$, Haibo Feng$^{2}$, Jun Hu$^{2}$, Jun Zhou$^{2}$, Xiaolu Zhang$^{2, \ddagger}$, Zhenzhong Lan$^{2, \ddagger}$, Junbo Zhao$^{2, 4, \ddagger}$, Da Zheng$^{2, \ddagger}$, Chongxuan Li$^{1, \dag}$, Jianguo Li$^{2, \dag}$, Ji-Rong Wen$^{1, \dag}$}

\footnotetext{$^\dag$Corresponding Authors. $^\ddagger$Team Leaders. $^*$Core Contributors.}

\affiliation{$^1$Renmin University of China, $^2$Ant Group, $^3$Shanghai Jiao Tong University, $^4$Zhejiang University}

\begin{document}
\maketitle

\input{doc/abstract}

\input{doc/intro}

\input{doc/relatedwork}

\input{doc/lladamoe}

\input{doc/experiments}

\input{doc/conclu}

\clearpage
\newpage

\bibliographystyle{antgroup}
\bibliography{ref/reference}

\end{document}

%% file: math_commands.tex
\usepackage{amsmath,amsfonts,bm}

\def\eqref#1{equation~\ref{#1}}

\def\1{\bm{1}}

\DeclareMathAlphabet{\mathsfit}{\encodingdefault}{\sfdefault}{m}{sl}
\SetMathAlphabet{\mathsfit}{bold}{\encodingdefault}{\sfdefault}{bx}{n}



%% file: setting-ai.tex
\newcommand*\justify{%
  \fontdimen2\font=0.4em
  \fontdimen3\font=0.2em
  \fontdimen4\font=0.1em
  \fontdimen7\font=0.1em
  \hyphenchar\font=`\-
}

\renewcommand{\texttt}[1]{%
  \begingroup
  \ttfamily
  \begingroup\lccode`~=`/\lowercase{\endgroup\def~}{/\discretionary{}{}{}}%
  \begingroup\lccode`~=`[\lowercase{\endgroup\def~}{[\discretionary{}{}{}}%
  \begingroup\lccode`~=`.\lowercase{\endgroup\def~}{.\discretionary{}{}{}}%
  \catcode`/=\active\catcode`[=\active\catcode`.=\active
  \justify\scantokens{#1\noexpand}%
  \endgroup
}

\usepackage{amsmath}
\usepackage{amssymb}
\usepackage{amsfonts}                               
\usepackage{amsthm}
\usepackage[mathcal]{eucal}
\usepackage{mathrsfs}
\usepackage{bm}                                     
\usepackage{blkarray}                               
\usepackage{nicefrac}                               

\usepackage{wrapfig}
\usepackage{graphicx}                               
\usepackage{caption}
\usepackage{cleveref}

\usepackage{tikz}                                          
\usepackage{circuitikz}
\usetikzlibrary{patterns,snakes}
\usetikzlibrary{positioning,calc,fit,decorations.pathmorphing,shapes.geometric, shapes.gates.logic.US, calc}
\usetikzlibrary{arrows,arrows.meta,decorations.markings,shapes,shapes.arrows}
\usetikzlibrary{decorations,decorations.pathreplacing}
\usetikzlibrary{backgrounds}
\usepackage{filecontents}                           
\usepackage{pgfplots}
\usepackage{pgfplotstable}
\usepgfplotslibrary{groupplots}
\usepackage{scalefnt}
\pgfplotsset{compat=newest}

%% file: doc/abstract.tex
\begin{abstract}
We introduce LLaDA-MoE, a large language diffusion model with the Mixture-of-Experts (MoE) architecture, trained from scratch on approximately 20T tokens. LLaDA-MoE achieves competitive performance with significantly reduced computational overhead by maintaining a 7B-parameter capacity while activating only 1.4B parameters during inference. Our empirical evaluation reveals that LLaDA-MoE achieves state-of-the-art performance among diffusion language models with larger parameters, surpassing previous diffusion language models LLaDA, LLaDA 1.5, and Dream across multiple benchmarks. The instruct-tuned model LLaDA-MoE-7B-A1B-Instruct demonstrates capabilities comparable to Qwen2.5-3B-Instruct in knowledge understanding, code generation, mathematical reasoning, agent and alignment tasks, despite using fewer active parameters. Our results show that integrating a sparse MoE architecture into the training objective of masked diffusion language models still brings out MoE's strengths under efficient inference with few active parameters, and opens ample room for further exploration of diffusion language models. LLaDA-MoE models are available at Huggingface\footnotemark{$^1$}\footnote{$^1$~https://huggingface.co/collections/inclusionAI/llada-68c141bca386b06b599cfe45}.
\end{abstract}

\begin{figure}[h]
  \centering
  \includegraphics[width=0.92\textwidth]{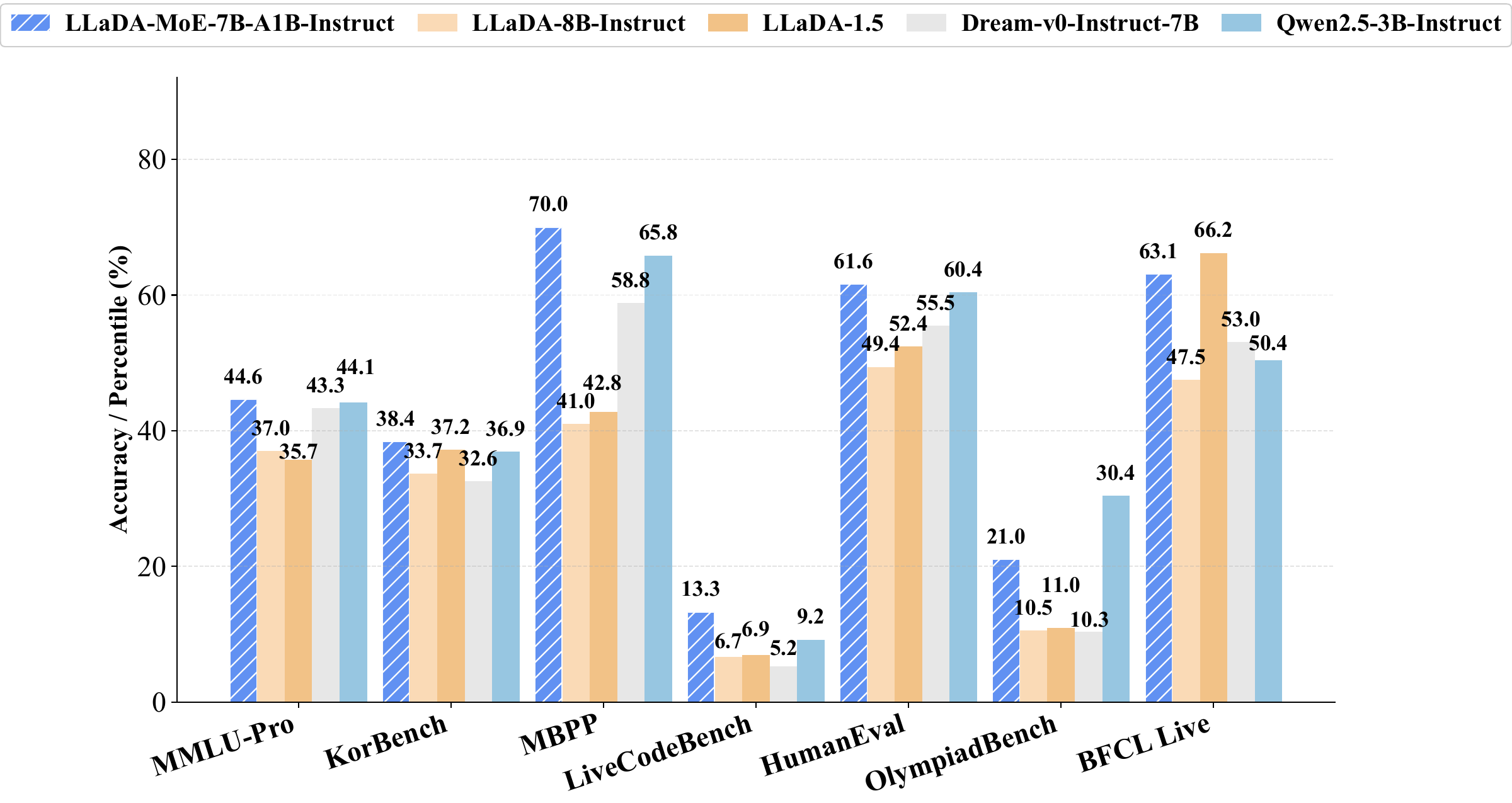}
  \caption{\textbf{Benchmark results.} We compare LLaDA-MoE with larger MDMs and Qwen2.5-3B-Instruct across key tasks in knowledge, reasoning, mathematics, coding, and agent tasks. Despite using fewer activated parameters, LLaDA-MoE consistently outperforms other diffusion language models and achieves performance comparable to Qwen2.5-3B-Instruct.}
  \label{fig:llada_moe_benchmark}
\end{figure}

%% file: doc/intro.tex
\section{Introduction}

Large Language Models (LLMs)~\citep{zhao2023survey} have advanced rapidly and are now widely used across a wide range of tasks. Alongside the dominant autoregressive (AR) paradigm~\citep{radford2018improving, radford2019language, brown2020language, ouyang2022training}, Masked Diffusion Models (MDMs)~\citep{austin2021structured, lou2023discrete, ou2024your, nie2025large} provide an alternative modeling paradigm with comparable scaling properties and performance~\citep{lou2023discrete, ou2024your, nie2024scaling, gong2024scaling, nie2025large}. These trends make diffusion language modeling a promising direction.

Despite these achievements, work on MDMs has largely relied on dense Transformer backbones~\citep{vaswani2017attention}, and, to the best of our knowledge, no prior work has pretrained MDMs from scratch with a sparse MoE architecture~\citep{jacobs1991adaptive, shazeer2017outrageously, lepikhin2020gshard, fedus2022switch, shen2023mixture, jiang2024mixtral}. By contrast, the MoE architecture has been widely validated for AR models~\citep{liu2024deepseekv2, liu2024deepseek, comanici2025gemini, bai2023qwen, yang2025qwen3}, achieving performance comparable to larger dense models while activating only a small subset of parameters per token via sparse expert routing. These observations provide a basis for studying sparse MoE architectures for MDMs.

In this work, we introduce LLaDA-MoE, a diffusion language model with a sparse MoE architecture that maintains a small number of active parameters during inference while delivering strong overall performance. With only 1B active parameters, LLaDA-MoE surpasses prior dense 8B diffusion language models~\citep{nie2025large, zhu2025llada, ye2025dream}. After instruction tuning, it achieves performance comparable to Qwen2.5-3B-Instruct~\citep{bai2023qwen} across knowledge understanding, code generation, mathematical reasoning, agent, and alignment tasks. Taken together, these results provide initial evidence that a sparse MoE architecture is a viable path toward more efficient MDMs.

\vspace{0.1cm}

We summarize our contributions as follows:

\begin{itemize}
    \item We introduce LLaDA-MoE, a diffusion language model with a sparse MoE architecture trained from scratch. It combines masked diffusion modeling with MoE to deliver strong performance while keeping the active-parameter budget small during inference.
    \item We demonstrate that LLaDA-MoE achieves state-of-the-art performance in diffusion language models: it surpasses prior 8B-parameter diffusion language models while activating only 1.4B parameters, and matches the performance of Qwen2.5-3B-Instruct across diverse tasks, including knowledge understanding, code generation, mathematical reasoning, agent, and alignment tasks.
\end{itemize}

%% file: doc/relatedwork.tex
\section{Related Work}

\subsection{Diffusion Language Models}

Recent studies investigate diffusion modeling for discrete data~\citep{austin2021structured, campbell2022continuous, chen2022analog, he2022diffusionbert, gong2022diffuseq, li2022diffusion, dieleman2022continuous, gulrajani2023likelihood}, with MDMs receiving increasing attention for simplicity and strong results. MDMs generate text by iteratively refining partially masked sequences, in contrast to widely used AR models that decode tokens from left to right~\citep{radford2018improving, radford2019language, brown2020language, ouyang2022training}. As illustrated in Figure~\ref{fig:llada_moe_overview}, an MDM repeatedly predicts the currently masked tokens and progressively unmasks them during inference. During training, MDMs optimize a variational lower bound on the log-likelihood by reconstructing masked tokens from partially observed context~\citep{lou2023discrete, ou2024your, sahoo2024simple, shi2024simplified}, thereby learning to recover the original sequence from incomplete inputs.

LLaDA~\citep{nie2025large} exemplifies the scaling behavior of MDMs~\citep{nie2024scaling}: trained from scratch with 8B parameters, it achieves performance comparable to LLaMA3 8B across a range of benchmarks~\citep{dubey2024llama}. Building on this, subsequent work has expanded the scope and practicality of diffusion language models by extending them to multimodal~\citep{you2025llada, yang2025mmada, li2025lavida, yu2025dimple}, improving inference efficiency through caching and parallel decoding strategies~\citep{wu2025fast, ma2025dkv, liu2025dllm, wei2025accelerating}, and strengthening reasoning and coding capabilities~\citep{zhu2025llada, gong2025diffucoder, zhao2025d1, tang2025wd1, huang2025reinforcing, wang2025revolutionizing}. These developments underscore the substantial potential of diffusion language models.

\subsection{The MoE Architecture} The MoE architecture improves parameter efficiency by activating only a small subset of parameters for each token: instead of a single dense feed-forward block, it maintains a large pool of expert networks and uses a learned router to select which experts to run per token~\citep{jacobs1991adaptive, shazeer2017outrageously, lepikhin2020gshard, fedus2022switch, shen2023mixture, jiang2024mixtral}. In a typical MoE layer, the router assigns each token to the top $k$ experts, and only those experts are activated. This directs computation to token-relevant parameters and allows the model to keep a much larger total parameter pool without activating all of it for every token. To prevent routing collapse and maintain balanced expert usage, MoE models typically add auxiliary load-balancing losses and enforce per-expert capacity limits. The integration of MDMs and MoE may provide the diffusion language modeling paradigm with a more resource-efficient and capacity-scalable framework.

%% file: doc/lladamoe.tex
\section{LLaDA-MoE}

\begin{figure}[t]
  \centering
  \includegraphics[width=\textwidth]{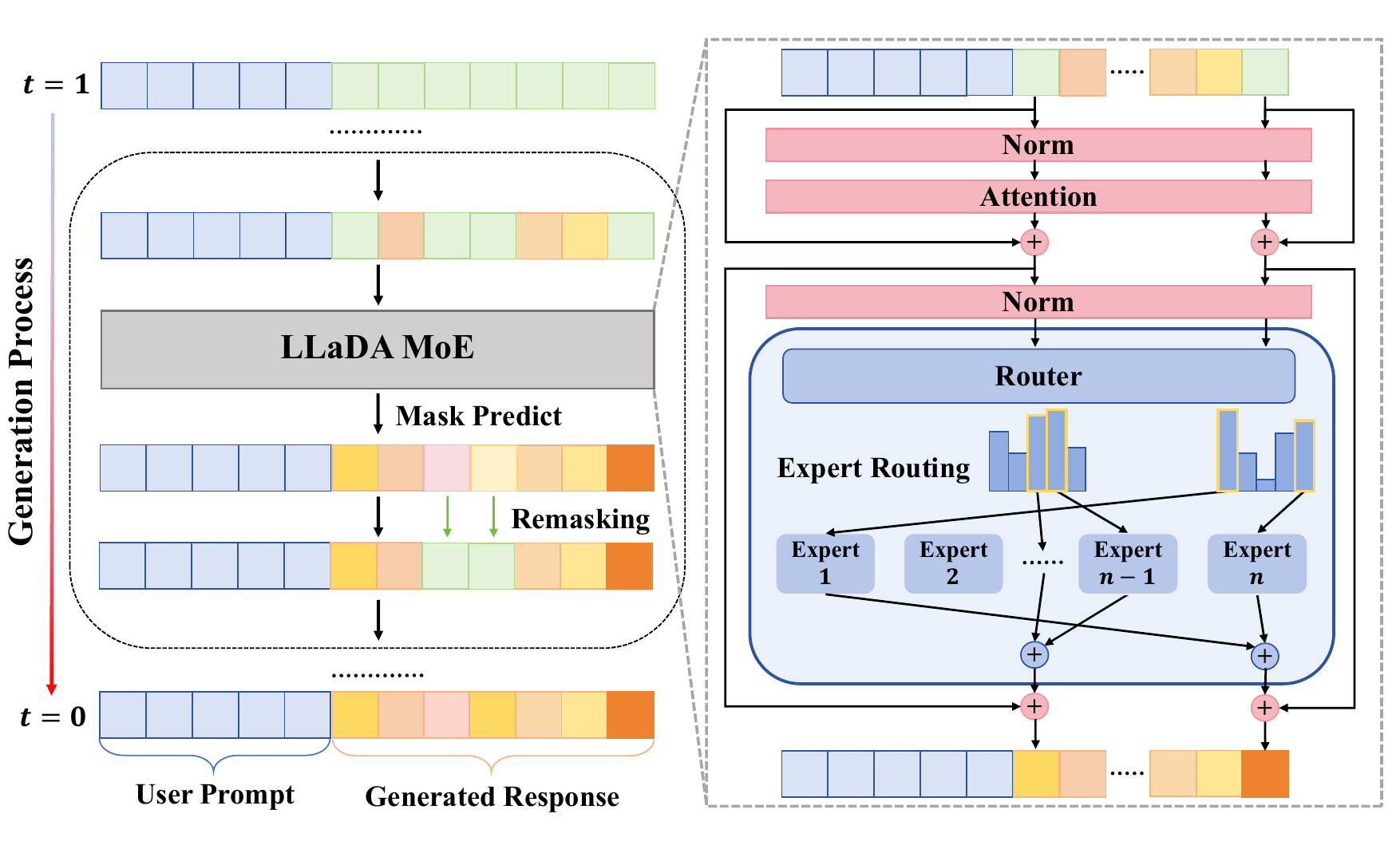}
  \caption{\textbf{Overview of the generation process and architecture.} \textbf{Left:} The iterative generation process from fully masked ($t=1$) to fully unmasked ($t=0$). \textcolor[HTML]{6C87BF}{Blue blocks} are fixed user prompt tokens, \textcolor[HTML]{588E32}{green blocks} are mask tokens. The model iteratively predicts and remasks tokens until generation completes. \textbf{Right:} The MoE architecture with router selecting top-$2$ experts per token. The histogram shows expert routing distribution, and outputs are weighted combinations of selected experts, enabling efficient sparse activation.}
  \label{fig:llada_moe_overview}
\end{figure}

\subsection{Overview}

\textbf{Model Architecture.} LLaDA-MoE employs RMSNorm for normalization, SwiGLU as the activation function, rotary positional embeddings for position encoding, and incorporates QK-layernorm within its multi-head attention blocks~\citep{zhang2019root, shazeer2020glu, su2024roformer}. Key architectural parameters are summarized in Table~\ref{tab:llada_moe_arch}.

\begin{table}[t]
\centering
\caption{Comparison among LLaDA‑MoE‑7B‑A1B‑Base with other MDMs and AR baselines.}
\label{tab:base_baseline}
\begin{adjustbox}{max width=\linewidth}
\begin{tabular}{lcccc}
\toprule
& \makecell{LLaDA-MoE-7B-A1B-Base}
& \makecell{LLaDA-8B-Base}
& \makecell{Dream-v0-Base-7B}
& \makecell{Qwen2.5-3B-Base} \\
\midrule
Architecture        & MoE           & Dense     & Dense                 & Dense     \\
Model               & Diffusion     & Diffusion & Diffusion             & AR        \\
Method              & Pretrain      & Pretrain  & Continue Pretrain     & Pretrain  \\
\# Total Params     & 7B            & 8B        & 7B                    & 3B        \\
\# Activated Params & 1B            & 8B        & 7B                    & 3B        \\
\midrule
\multicolumn{5}{c}{\textit{General Tasks}} \\
MMLU        & 64.59 & 65.90 & \textbf{69.50} & \underline{67.98} \\
MMLU-Pro    & 39.16 & \underline{41.80} & \textbf{48.15} & 35.50 \\
CEval       & 65.56 & \underline{70.50} & 59.18 & \textbf{75.00} \\
CMMLU       & 65.65 & \underline{69.90} & 60.87 & \textbf{73.65} \\
RACE        & 84.96 & \textbf{88.37} & 44.70 & \underline{87.88} \\
\midrule
\multicolumn{5}{c}{\textit{Reasoning Tasks}} \\
BBH         & 52.71 & 49.80 & \textbf{57.90} & \underline{56.50} \\
Drop        & 65.86 & \underline{72.93} & \textbf{75.16} & 51.61 \\
KorBench    & 31.20 & \underline{33.68} & \textbf{37.44} & 27.44 \\
\midrule
\multicolumn{5}{c}{\textit{Math Tasks}} \\
GSM8K           & 66.41 & 70.70 & \underline{77.79} & \textbf{78.17} \\
MATH            & 36.10 & 27.30 & \underline{39.60} & \textbf{40.94} \\
OlympiadBench   & \underline{10.07} & 6.85  & \textbf{10.22} & 9.33  \\
\midrule
\multicolumn{5}{c}{\textit{Coding Tasks}} \\
CRUX-O              & \textbf{39.00} & 31.00 & \underline{37.75} & 35.62 \\
MBPP                & 52.40 & 38.20 & \underline{56.20} & \textbf{69.56} \\
MultiPL-E           & \textbf{41.13} & 23.61 & 27.60 & \underline{40.80} \\
HumanEval           & 45.73 & 33.50 & \underline{57.90} & \textbf{57.93} \\
LiveCodeBench v6    & \underline{16.18} & 2.53  & 14.87 & \textbf{16.99} \\
BigCodeBench-Full   & \underline{21.23} & 13.42 & 18.33 & \textbf{30.88} \\
\midrule
Avg & \underline{46.94} & 43.53 & 46.66 & \textbf{50.34} \\
\bottomrule
\end{tabular}
\end{adjustbox}
\end{table}

\begin{figure}[t]
  \centering
  \includegraphics[width=\textwidth]{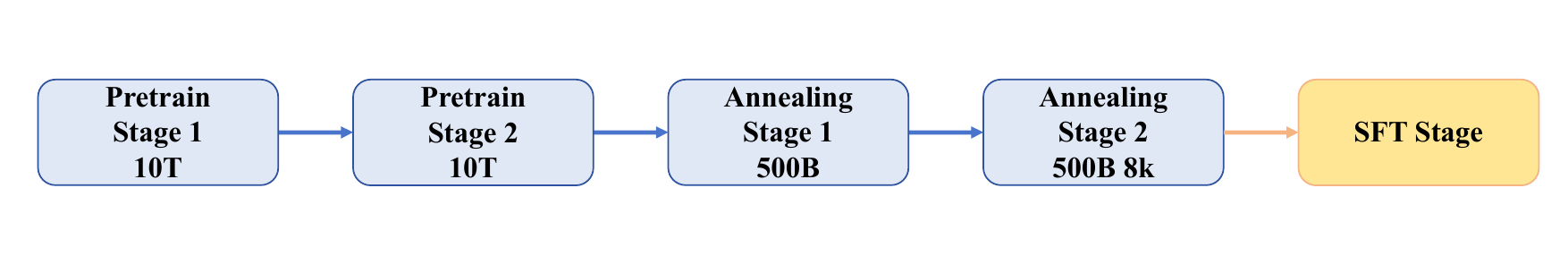}
  \caption{\textbf{Training pipeline.} LLaDA-MoE is trained through Pretrain stage 1 (10T tokens), pretrain stage 2 (10T tokens), annealing stage 1 (500B tokens), annealing stage 2 (500B tokens with 8k context length), followed by SFT on curated prompt–answer pairs.}
  \label{fig:train_pipeline}
\end{figure}

\textbf{Training pipeline.} As shown in Figure~\ref{fig:train_pipeline}, our training pipeline comprises multiple phases. Pretrain Stage 1 trains the model from scratch for 10T tokens on a large mixed text corpus. Pretrain Stage 2 then continues for another 10T tokens resampled from the same underlying corpus, with sampling reweighted to increase the fraction of mathematics and code. For Annealing Stage 1, we initialize from the checkpoint with the best average evaluation metrics from Pretrain Stage 2 and continue training on 500B tokens of high-quality text. Annealing Stage 2 resumes from the last Annealing Stage 1 checkpoint, raises the RoPE base from 10,000 to 50,000, expands the context length from 4k to 8k to support longer sequences, and trains for 500B tokens. Finally, the SFT Stage performs supervised tuning on high-quality question–answer pairs according to Eq.~\ref{eq:llada_sft}; the resulting checkpoint serves as our Instruct model.

\begin{table}[t]
    \centering
    \caption{The LLaDA-MoE Architecture}
    \label{tab:llada_moe_arch}
    \begin{tabular}{lr}
        \toprule
        Layers                              & 16 \\
        Hidden Dimension                    & 2048 \\
        Attention Heads                     & 16 \\
        Total Experts                       & 64 \\
        Activated Experts                   & 8 \\
        Expert Dimension                    & 1024 \\
        RoPE Base                           & 50,000 \\
        Active Parameters                   & 1.4B \\
        Non-embedding Parameters            & 7B \\
        \bottomrule
    \end{tabular}
\end{table}

\subsection{Training and Inference for LLaDA-MoE}

We follow the standard formulation of MDMs to present the modeling, training objective, and inference procedure, and then describe pretrain, supervised fine-tuning (SFT), and related methods~\citep{austin2021structured, lou2023discrete,sahoo2024simple,shi2024simplified,ou2024your,nie2025large}.

\textbf{Forward Process.} Let $y \in \{0, 1, \dots, K - 1 \}^L$ be a clean sequence of length $L$ over a vocabulary of size $K$. We first draw a noise level $t \sim \mathcal{U}[0, 1]$, then independently decide for each position whether to keep the original token or replace it by $\mathbf{M}$, where $\mathbf{M}$ denotes the mask token. Formally,
\begin{align}
    q(y_t \mid t, y) \;=\; \prod_{i=1}^{L} q(y_t^i \mid t, y^i), \qquad
    q(y_t^i \mid t, y^i) =
    \begin{cases}
        1-t, & y_t^i = y^i, \\
        t,   & y_t^i = \mathbf{M}, \\
        0,   & \text{otherwise}.
    \end{cases}
    \label{eq:llada_forward}
\end{align}

\textbf{Mask predictor and objective.} A parametric mask predictor $p_\theta(\cdot \mid y_t)$ outputs token distributions for all positions. The training objective of LLaDA-MoE is
\begin{align}
    \mathcal{L}_{\text{Pretrain}}(\theta) = 
    -\mathbb{E}_{y \sim p_{\text{data}}}
    \mathbb{E}_{t \sim \mathcal{U}[0,1]}
    \mathbb{E}_{y_t \sim q(y_t \mid t, y)}
    \left[
    \frac{1}{t} \sum_{i=1}^{L} \mathbf{1}[y_t^i=\mathbf{M}] \log p_\theta(y^i \mid y_t)
    \right],
    \label{eq:llada_pretrain}
\end{align}
$\mathcal{L}_{\text{Pretrain}}$ upper-bounds the negative log-likelihood of the model distribution.

LLaDA‑MoE employs bidirectional attention and is pretrained with a fixed 4k context, creating a train–test discrepancy: training always sees 4k tokens, whereas inference contexts vary and are often shorter, which degrades performance. To narrow this gap, we adapt variable-length training~\citep{nie2024scaling} for a small fraction of steps. Specifically, during pretraining, in $1\%$ of steps we sample a target length $\ell \in [8, 4096]$ and truncate the input to $\ell$ tokens; the remaining $99\%$ of steps use the default 4k context. This simple intervention reduces the distribution mismatch and yields substantial gains on evaluation metrics.

\textbf{MoE Routing.} We employ a top‑$k$ gated MoE layer:
\begin{align}
    p_t = \mathrm{Softmax}(\mathrm{Router}(h_t)), \qquad
    o_t = \sum_{i} p_{t,i}\, E_i(h_t), \qquad \text{where } p_{t,i} \in \mathrm{Topk}(p_t).
\end{align}
where $h_t$ is the hidden state, $\mathrm{Router}(\cdot)$ is a linear router producing per‑expert logits, $E_i(\cdot)$ is the $i$-th expert, and $\mathrm{Topk}(\cdot)$ selects the $k$ largest entries; the corresponding softmax scores $p_{t,i}$ weight the selected experts’ outputs.

\begin{figure}[t]
    \centering
    \begin{subfigure}{0.42\linewidth}
        \centering
        \includegraphics[width=\linewidth]{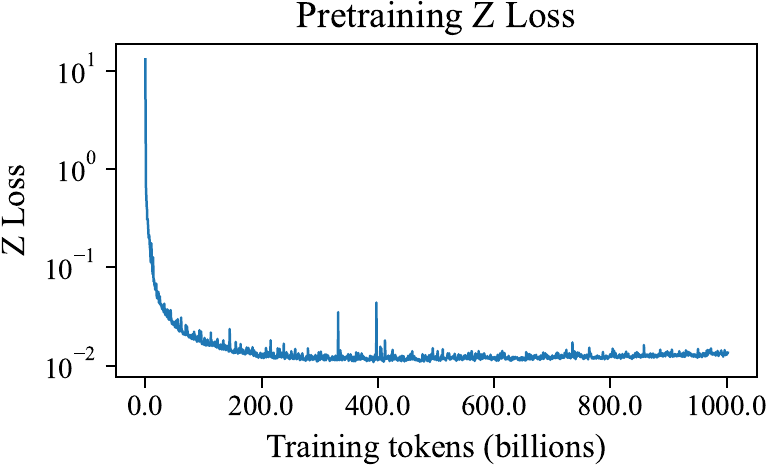}
    \end{subfigure}
    \hspace{0.02\linewidth}
    \begin{subfigure}{0.42\linewidth}
        \centering
        \includegraphics[width=\linewidth]{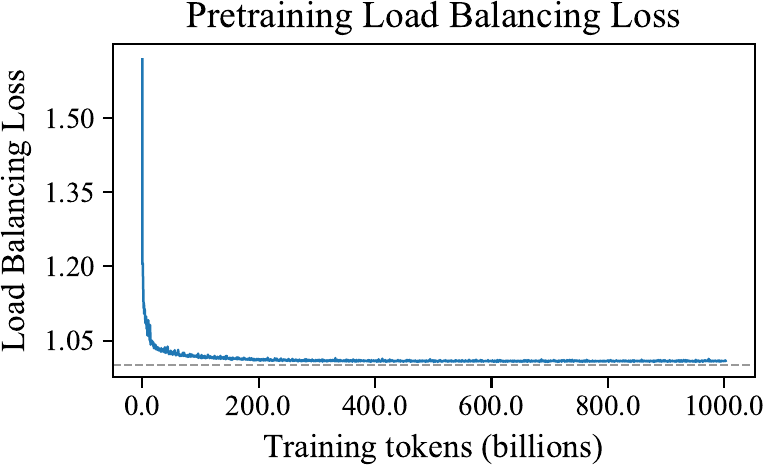}
    \end{subfigure}
    \caption{\textbf{Training dynamics of auxiliary losses over training tokens.} LLaDA‑MoE pre‑training results over the first 1T tokens. Left: Z‑Loss; right: Load‑Balancing Loss.}
    \label{fig:aux_losses}
\end{figure}

To mitigate expert‑load imbalance, we adopt the standard auxiliary losses~\citep{shazeer2017outrageously, zoph2022st}:
\begin{align}
\mathcal{L}_{\mathrm{LB}} = N \sum_{i=1}^{N} f_i\, P_i, \qquad
\mathcal{L}_{\mathrm{Z}} = \frac{1}{T}\sum_{t=1}^{T}\left(\log \sum_{j=1}^{N} e^{z_{t,j}}\right)^2,
\end{align}
where $z_t = \mathrm{Router}(h_t)$, $p_t = \mathrm{Softmax}(z_t)$, $P_i$ denotes the average routing probability assigned to expert $i$ across tokens, $f_i$ denotes the frequency of expert $i$ being selected across all tokens, $N$ is the number of experts, and $T$ is the number of tokens. In practice, we set loss weights of 0.01 for $\mathcal{L}_{\mathrm{LB}}$ and 0.001 for $\mathcal{L}_{\mathrm{Z}}$, which yields stable expert‑routing training. Figure~\ref{fig:aux_losses} shows the training dynamics of LLaDA‑MoE’s auxiliary losses over the first 1T training tokens, where both the Z‑Loss and the load‑balancing loss decrease rapidly early in pre‑training and then stabilize at low magnitudes.

\textbf{Supervised Fine-Tuning.} SFT is a special case of the pretraining objective: we apply the corruption kernel in Eq.~\ref{eq:llada_forward} only to the response $y$ while keeping the prompt $x$ clean. Given a pair $(x, y)$, sample $t \sim \mathcal{U}[0,1]$, form $y_t$ by masking each token in $y$ with probability $t$, and train the model to recover masked tokens in $y$ conditioned on $x$:
\begin{align}
    \mathcal{L}_{\text{SFT}}(\theta)
    &=
    -\mathbb{E}_{(x, y)\sim p_{\text{data}}}
    \mathbb{E}_{t \sim \mathcal{U}[0, 1]}
    \mathbb{E}_{y_t \sim q(y_t \mid t, y)}
    \left[
    \frac{1}{t} \sum_{i = 1}^{|y|}
    \mathbf{1}[y_t^i = \mathbf{M}] \log p_\theta(y^i \mid x, y_t)
    \right].
    \label{eq:llada_sft}
\end{align}

To enable variable-length generation, we pad shorter responses with $|\text{EOS}|$ tokens to the maximum length in the batch, consistent with LLaDA~\citep{nie2025large}. The appended $|\text{EOS}|$ tokens are treated as part of the response: they are masked and included in the loss in Eq.~\ref{eq:llada_sft}. For multi-turn dialogs $D=[(x_1, y_1), \ldots, (x_T, y_T)]$, we sample a target turn $\tau \in [1, T]$. The visible prompt is $[x_1, y_1, \ldots, x_\tau]$. We apply the corruption kernel in Eq.~\ref{eq:llada_forward} only to the target response $y_\tau$ to obtain $y_{\tau,t}$. We then train using Eq.~\ref{eq:llada_sft} to recover the masked tokens in $y_\tau$ given the visible prompt and $y_{\tau,t}$. Any $|\text{EOS}|$ tokens appended after $y_\tau$ are included in the target segment and handled in the same way.

As noted previously, LLaDA-MoE’s context window is expanded from 4k to 8k during Annealing Stage 2. However, during SFT, we limit each sample to 4k tokens because most SFT samples are shorter than 4k; masking and training on sequences padded to 8k with $|\text{EOS}|$ token may cause the model to generate too many $|\text{EOS}|$ tokens and degrade performance.

\textbf{Inference.} Starting from the fully masked sequence $\mathbf{M}^L$, we iteratively reduce the noise level and sample tokens at masked positions using the mask predictor. For $0 \le s < t \le 1$,
\begin{align}
    q(y_s \mid s, t, y_t) = \prod_{i = 1}^{L} q(y_s^i \mid s, t, y_t), \qquad q(y_s^i \mid s, t, y_t) =
    \begin{cases}
        \frac{t-s}{t} p_\theta(y^i \mid y_t), & y_t^i=\mathbf{M}, y_s^i \neq \mathbf{M}, \\
        \frac{s}{t}, & y_t^i=\mathbf{M}, y_s^i=\mathbf{M}, \\
        1, & y_t^i \neq \mathbf{M}, y_s^i = y_t^i, \\
        0, & \text{otherwise}.
    \end{cases}
    \label{eq:llada_reverse}
\end{align}

Beyond the sampling scheme above, prior work adopts a semi‑autoregressive sampling strategy~\citep{arriola2025block, nie2025large}. Specifically, a sequence of length $L$ is partitioned into $K=L/B$ blocks, each containing $B$ tokens. At block $k$, all $B$ masked positions are predicted in parallel using the same reverse dynamics as above; once block $k$ is fully unmasked, decoding proceeds to block $k+1$. In other words, autoregression operates at the block level. Semi‑autoregressive sampling is often combined with low‑confidence remasking. At each decoding step, we record the chosen token’s probability as its confidence and then remask the tokens with the lowest confidence to refine the output results.

\begin{table}[t]
    \centering
    \caption{Comparison among LLaDA‑MoE‑7B‑A1B‑Instruct with other MDMs and AR baselines.}
    \label{tab:instruct_baseline}
    \begin{adjustbox}{max width=\linewidth}
    \begin{tabular}{lccccc}
    \toprule
    & \makecell{LLaDA-MoE-7B-A1B-Instruct}
    & \makecell{LLaDA-8B-Instruct}
    & \makecell{LLaDA-1.5}
    & \makecell{Dream-v0-Instruct-7B}
    & \makecell{Qwen2.5-3B-Instruct} \\
    \midrule
    Architecture        & MoE               & Dense             & Dense                 & Dense                     & Dense                 \\
    Model               & Diffusion         & Diffusion         & Diffusion             & Diffusion                 & AR                    \\
    Method              & Pretrain + SFT    & Pretrain + SFT    & Pretrain + SFT + DPO  & Continue Pretrain + SFT   & Pretrain + SFT + RL   \\
    \# Total Params     & 7B                & 8B                & 8B                    & 7B                        & 3B                    \\
    \# Activated Params & 1B                & 8B                & 8B                    & 7B                        & 3B                    \\
    \midrule
    \multicolumn{6}{c}{\textit{General Tasks}} \\
    MMLU        & \underline{67.18} & 65.50 & 66.00 & 67.00 & \textbf{69.11} \\
    MMLU-Pro    & \textbf{44.64} & 37.00 & 35.70 & 43.30 & \underline{44.13} \\
    CMMLU       & \underline{64.30} & 55.21 & 58.72 & 58.82 & \textbf{65.62} \\
    CEval       & \underline{63.93} & 54.48 & 58.41 & 57.98 & \textbf{68.20} \\
    
    \midrule
    \multicolumn{6}{c}{\textit{Reasoning Tasks}} \\
    Drop        & 79.77 & \underline{83.09} & \textbf{84.89} & 76.25 & 68.56 \\
    KorBench    & \textbf{38.40} & 33.68 & \underline{37.20} & 32.56 & 36.88 \\
    \midrule
    \multicolumn{6}{c}{\textit{Math Tasks}} \\
    GSM8K           & 82.41 & 78.60 & \underline{83.30} & 81.00 & \textbf{86.28} \\
    MATH            & \underline{58.68} & 42.20 & 42.60 & 39.20 & \textbf{67.02} \\
    OlympiadBench   & \underline{21.04} & 10.52 & 10.96 & 10.44 & \textbf{30.41} \\
    \midrule
    \multicolumn{6}{c}{\textit{Coding Tasks}} \\
    CRUX-O              & \underline{42.38} & 28.50 & 29.12 & 40.12 & \textbf{46.75} \\
    MBPP                & \textbf{70.02} & 41.00 & 42.80 & 58.80 & \underline{65.81} \\
    MultiPL-E           & \underline{52.53} & 29.08 & 29.04 & 29.86 & \textbf{54.92} \\
    HumanEval           & \textbf{61.59} & 49.40 & 52.40 & 55.50 & \underline{60.37} \\
    LiveCodeBench v6    & \textbf{13.27} & 6.66  & 6.94  & 5.23  & \underline{9.20} \\
    BigCodeBench-Full   & \underline{20.44} & 11.32 & 11.93 & 19.04 & \textbf{27.81} \\
    \midrule
    \multicolumn{6}{c}{\textit{Agent \& Alignment Tasks}} \\
    IFEval Strict Prompt    & \underline{59.33} & 51.39 & 58.23 & \textbf{62.50} & 58.20 \\
    BFCL-Live               & \underline{63.09} & 47.47 & \textbf{66.20} & 53.03 & 50.40 \\
    \midrule
    Avg & \underline{53.12} & 42.65 & 45.56 & 46.51 & \textbf{53.51} \\
    \bottomrule
    \end{tabular}
    \end{adjustbox}
\end{table}

%% file: doc/experiments.tex
\section{Experiments}

We evaluate LLaDA‑MoE on a broad suite of benchmarks spanning knowledge~\citep{hendrycks2020measuring, wang2024mmlu, li2023cmmlu, huang2023c, lai2017race}, reasoning~\citep{suzgun2022challenging, dua2019drop, ma2024kor}, mathematics~\citep{cobbe2021training, hendrycks2021measuring, he2024olympiadbench}, coding~\citep{gu2024cruxeval, austin2021program, cassano2022multipl, chen2021evaluating, jain2024livecodebench, zhuo2024bigcodebench}, agent~\citep{, patilberkeley}, and alignment~\citep{zhou2023instruction}.

For MDMs~\citep{nie2025large, zhu2025llada, ye2025dream} with reported results, we cite the published numbers directly. For all other generative benchmarks, we use semi‑autoregressive sampling~\citep{arriola2025block, nie2025large} with a generation length of 1024 and a block length of 64 to ensure a fair and robust comparison.

Table~\ref{tab:base_baseline} and Table~\ref{tab:instruct_baseline} report the evaluation results for LLaDA‑MoE‑Base and LLaDA‑MoE‑Instruct. Overall, despite activating only 1B parameters, LLaDA‑MoE‑7B‑A1B is highly competitive. On average, both the Base and Instruct outperform the previous larger dense MDM baselines—LLaDA‑8B and Dream‑7B—as well as LLaDA‑1.5. After instruction tuning, LLaDA‑MoE‑7B‑A1B‑Instruct trails Qwen2.5‑3B‑Instruct by only a small margin.

In summary, LLaDA‑MoE shows that masked diffusion language modeling works well with a sparse MoE architecture, delivering strong and parameter‑efficient performance and opening a broad design space for future exploration.

%% file: doc/conclu.tex
\section{Conclusion}

We introduce LLaDA-MoE, a diffusion language model trained from scratch with a MoE architecture enabling efficient inference. With 1.4B active parameters, LLaDA-MoE surpasses prior 8B-parameter diffusion language models. After instruction tuning, it is comparable to Qwen2.5-3B-Instruct across all benchmarks. These results establish MoE as an effective foundation for efficient MDMs and open the door to further research and improvements.

This study is potentially constrained by the current model size; in future work, we plan to scale LLaDA-MoE and address any challenges that accompany scaling up.